\documentclass[letterpaper]{article}

\usepackage{cite, graphicx, amsmath, amsfonts, amssymb, array, latexsym, url}
\usepackage{multirow, dsfont, booktabs}

\newcommand{\Aa}{\mathcal{A}}
\newcommand{\Hh}{\mathcal{H}}
\newcommand{\argmax}{\operatornamewithlimits{argmax}}
\newcommand{\argmin}{\operatornamewithlimits{argmin}}

\begin{document}

\title{Regularized Richardson-Lucy Algorithm for Sparse Reconstruction of Poissonian Images}

\author{Elad~Shaked and Oleg~Michailovich%
\thanks{E. Shaked and O. Michailovich are with the School of Electrical and Computer Engineering, University of Waterloo, Canada N2L 3G1 (phone: 519-888-4567; e-mails: \{eshaked, olegm\}@uwaterloo.ca).}}

\maketitle

\begin{abstract}
Restoration of digital images from their degraded measurements has always been a problem of great theoretical and practical importance in numerous applications of imaging sciences. A specific solution to the problem of image restoration is generally determined by the nature of degradation phenomenon as well as by the statistical properties of measurement noises. The present study is concerned with the case in which the images of interest are corrupted by convolutional blurs and Poisson noises. To deal with such problems, there exists a range of solution methods which are based on the principles originating from the fixed-point algorithm of Richardson and Lucy (RL). In this paper, we provide conceptual and experimental proof that such methods tend to converge to sparse solutions, which makes them applicable only to those images which can be represented by a relatively small number of non-zero samples in the spatial domain. Unfortunately, the set of such images is relatively small, which restricts the applicability of RL-type methods. On the other hand, virtually all practical images admit sparse representations in the domain of a properly designed linear transform. To take advantage of this fact, it is therefore tempting to modify the RL algorithm so as to make it recover  representation coefficients, rather than the values of their associated image. Such modification is introduced in this paper. Apart from the generality of its assumptions, the proposed method is also superior to many established reconstruction approaches in terms of estimation accuracy and computational complexity. This and other conclusions of this study are validated through a series of numerical experiments.

\end{abstract}

%--------------------------------------------------- Introduction -----------------------------------------------------
\section{Introduction}\label{Intro}
The notion of {\em event counts} is fundamental in many imaging modalities. Thus, for instance, this notion is used to quantify the reception of gamma photons in nuclear imaging \cite{Shepp82, Lee95, Yavuz98, Bauschke99} as well as to describe the formation of optical images in charge coupled devices (CCD) \cite{Boie92, Healey94}. Confocal microscopy \cite{Holmes92}, astronomical \cite{Bradt04} and turbulent imaging \cite{Roggeman96} are additional examples of applications in which the notion of even counts routinely arises. In all these cases, working with event counts entails using a specific statistical assumption on the nature of acquired images. In particular, the images are assumed to be formed as blurred versions of their original counterparts contaminated by Poisson noise. Consequently, to recover the original images, the combined effect of the blur and noise needs to be reversed. A novel approach to the solution of this classical inverse problem is proposed in this work.

Since in practice, digital images are discrete and finite-dimensional objects, it seems reasonable to represent the original image $f$ along with its measured version $g$ by $N \times M$ real-valued matrices. Moreover, since in Poissonian imaging the values of both $f$ and $g$ have the interpretation of event counts, the images can be further constrained to be members of the subspace $\mathbb{V}$ of nonnegative-valued matrices. Formally,
\begin{equation}\label{sspace}
f, \, g \, \in \, \mathbb{V} = \left\{ y \in \mathbb{R}^{N \times M} \mid y_{n,m} \geq 0, n=0,1,\ldots,N-1, m=0,1,\ldots, M-1 \right\}.
\end{equation} 
Additionally, let $\mathcal{H}: \mathbb{V} \rightarrow \mathbb{V}$ be a linear operator describing the convolution of $f$ with a non-negative mask $h \geq 0$ of some size. Then, the formation of $g$ can be formally expressed as
\begin{equation}\label{model}
g = \mathcal{P} \left\{ \mathcal{H}\{f\} \right\},
\end{equation}
where $\mathcal{P}$ describes the effect of contamination of $\Hh\{f\}$ by Poisson noise. Therefore, as was mentioned before, to recover the original image $f$ from $g$, the combined effect of $\mathcal{P}\{{\mathcal{H}} \{\cdot\}\}$ in (\ref{model}) needs to be inverted.

The current arsenal of image restoration approaches to the solution of (\ref{model}) is relatively broad. In general, all these methods can be divided into two main groups, where the methods of the first group assume $\Hh$ to be identity, whereas the methods of the second group allow it to be a general low-pass filter. Within the first group, a range of available reconstruction methods take advantage of a {\em variance stabilized transformation} (VST) \cite{Anscombe48}, which allows the Poisson noise in (\ref{model}) to be transformed into approximately white Gaussian noise. Some recent developments in this direction include the works in \cite{Fryzlewicz04, Jansen06, Zhang08}, in which wavelet de-noising has been exploited to reject the ``gaussianized" Poisson noise. Conceptually similar ideas have been also advocated in \cite{Kolaczyk00, Zhang07} based on more advanced tools of statistical analysis. The framework of Bayesian estimation was exploited in \cite{Kolaczyk99, Nowak99, Timmermann99, Nowak00, Sardy04}, where hierarchical (Markovian) models have been used to describe a priori probabilities of the original images. Although all the above methods can be used to recover an approximation of $f$ in (\ref{model}), their applicability is restricted to the case of weak blurs, where $\Hh$ can be closely approximated by the identity operator ${\rm Id}$. Moreover, all these methods depend on VST, whose performance is known to deteriorate considerably in low-count settings~\cite{Antoniadis01}.

One of the nowadays classical methods of solving (\ref{model}) for the case of a general $\Hh$ was proposed in the beginning of the 1970s in the seminal works of W.~Richardson \cite{Richardson72} and L.~Luci \cite{Lucy74}. This method -- known as the Richardson-Lucy (RL) algorithm -- can be classified as a {\it maximum likelihood} (ML) estimator. As a general rule, however, ML estimation may result in less accurate and/or stable reconstructions as compared with {\em maximum a posteriori} (MAP) estimation methods based on the Bayesian paradigm. Consequently, the RL algorithm has been recently extended under the MAP estimation framework, resulting in a number of regularized solutions to (\ref{model}) which differ in the way the original image $f$ is modelled. Thus, for example, Gaussian models have been exploited in \cite{Molina94, Kempen00}, while the algorithm in \cite{Dey06} is based on the total variation model of \cite{Rudin92}. Unfortunately, there are conditions on which the above methods can result in erroneous reconstructions (as will be demonstrated later in the paper).

Despite the relative simplicity of the RL algorithm, it still remains one of the most widespread methods used in the current practice of Poissonian imaging. The main advantage of the RL algorithm stems from its nonlinear nature, allowing one to recover the high-frequency components of the original images which are the most affected by blur. Moreover, a closer look at the analytical properties of the RL algorithm reveals the fact that it has a ``built-in" ability to converge to {\em sparse} solutions, and, as a result, the application of the algorithm for the restoration of sparse images should be expected to produce particularly useful reconstructions. Unfortunately, most of the real-life images are not sparse in the spatial domain, in which case the RL algorithm may fail to provide useful results. On the other hand, most of such images {\em are} sparse in the domain of a properly designed linear transform \cite{Donoho95, Chen01, Aharon06, Elad07}. Consequently, to extend the applicability of the RL algorithm to a wider range of imagery data, it is tempting to find a way to apply the algorithm in the transformed domain, as opposed to the spatial domain. Accordingly, the main contribution of this paper consists in the introduction of a novel approach to the solution of (\ref{model}) which exploits the above idea. Moreover, it will be shown via an experimental study that the proposed method outperforms a number of alternative algorithms in terms of normalized mean-square error (NMSE), SSIM quality index \cite{Wang04}, as well as its stability and computational efficiency.

The rest of the paper is organized as follows. Section II provides some necessary details on the RL algorithm. The statistical models and assumptions underpinning the proposed method are detailed in Section III, while Section IV summarizes the main structure of the proposed solution method. The results of both computed-simulated and real-life experiments are presented in Section V. Section VI finalizes the paper with a discussion and conclusions.

\section{Richardson-Lucy Approach}
To establish a necessary foundation for the development and presentation of the proposed method, a brief overview of the RL algorithm is provided first. Under the image formation model of (\ref{model}), the likelihood function of the measured image $g$ can be defined as
\begin{equation}
P(g \mid f) = \prod_{n=0}^{N-1} \prod_{m=0}^{M-1} \frac{e^{-(\Hh\{f\})_{n,m}}  (\Hh\{f\})_{n,m}^{g_{n,m}} }{g_{n,m}!},
\end{equation}
where the samples $g_{n,m}$ of $g$ are assumed to be mutually independent. Consequently, defining $E(f) = - \log P(g \mid f)$, the ML estimate $\hat{f}_{ML}$ of $f$ is given by
\begin{equation}\label{RLcosta}
\hat{f}_{ML} = \argmin_{f \in \mathbb{V}} \{ E(f) \},
\end{equation}
where
\begin{equation}\label{RLcostb}
E(f) = \langle {\bf 1}, \mathcal{H}\{f\} \rangle - \langle g, \log(\mathcal{H}\{f\})  \rangle ,
\end{equation}
with $\langle \cdot , \cdot \rangle$ denoting the standard inner product in $\mathbb{R}^{N\times M}$. Differentiating $E(f)$ with respect to $f$ and equating the derivative to zero results in
\begin{equation}
\Hh^\ast \left\{ {\bf 1} \right\} = \Hh^\ast \left\{ \frac{g}{\Hh\{f\}} \right\},
\end{equation}
where $\bf 1$ denotes an $N \times M$ matrix of ones, the fractional line stands for a point-wise division, and $\Hh^\ast: \mathbb{V} \rightarrow \mathbb{V}$ denotes the adjoint of $\Hh$. Note that if $\Hh$ represents the operation of 2-D convolution with a positive-valued kernel $h > 0$, then $\Hh^\ast$ represents the convolution with a spatially reversed version of the same kernel $h$. Moreover, if $h$ is normalized to satisfy $\sum_{n,m} h_{n,m} = 1$, then both $\Hh\{{\bf 1}\}$ and $\Hh^\ast\{{\bf 1}\}$ are obviously equal to $\bf 1$. This fact is exploited by the RL algorithm, which recovers an approximation of the original image $f$ as a stationary point of the sequence of solutions produced by
\begin{equation}\label{RL}
f^{(t+1)} = f^{(t)} \cdot \mathcal{H}^\ast \left\{ \frac{g}{ \mathcal{H}\{ f^{(t)} \} } \right\},
\end{equation}
where the dot stands for a point-wise multiplication, and $t$ is the iteration index.

To better understand the nature of the solution produced by the minimization of (\ref{RLcostb}), consider the following steps. Using the definition of an adjoint operator, the cost functional $E(f)$ in (\ref{RLcostb}) can be rewritten as
\begin{align}
E(f) &= \langle \mathcal{H}^\ast \{{\bf 1}\} , f \rangle - \langle g, \log(\mathcal{H}\{f\})  \rangle =  \\
&= \langle {\bf 1} , f \rangle - \langle g, \log(\mathcal{H}\{f\})  \rangle \nonumber,
\end{align}
Moreover, since $f$ is assumed to be positive valued, the inner product $\langle {\bf 1} , f \rangle$ coincides with the $\ell_1$-norm of $f$ defined as $\|f\|_1 = \sum_{n,m}{|f_{n,m}|}$. As a result, the cost function $E(f)$ in (\ref{RLcosta}) can be further redefined as 
\begin{equation}\label{CostL1}
E(f)= \|f\|_1 - \langle g, \log(\mathcal{H}\{f\})  \rangle.
\end{equation}

A closer look into the two terms in (\ref{CostL1}) leads to a number of remarkable observations. First and foremost, the presence of $\ell_1$-norm in the cost function implies that the solution of (\ref{RLcosta}) will tend to be sparse \cite{Donoho95, Chen01, Elad06, Donoho06, Candes06}. {\it Therefore, the RL algorithm has a ``built-in" ability to converge to sparse solutions.} Moreover, such solutions are guaranteed to be positive due to the second term in (\ref{CostL1}) which works akin to a logarithmic barrier \cite[Ch.11]{Boyd04}. This is what makes the RL algorithm to favour the reconstructions which are sparse in the spatial domain.   

It goes without saying that the assumption of spatially sparse images may be unacceptable for a variety of natural scenes. For this reason, the RL algorithm is known to perform reliably in the case of, e.g., astronomical images \cite{Richardson72}, while its application to piecewise smooth images can produce rather disappointing results \cite{Dey06}. On the other hand, sparsity is known to be an extremely useful and liable assumption to use for describing the behaviour of the representation coefficients of natural images in the domain of certain linear transforms. Therefore, to extend the applicability of the RL algorithm to the above case, it is imperative to find a way to apply the algorithm directly to the (sparse) representation coefficients, rather than to their corresponding images. A possible variant of such an approach is detailed in the section that follows.

Before turning to the description of our method, a number of important properties of the RL recursion in (\ref{RL}) are worth to be mentioned. In particular, it can be seen from (\ref{RL}) that the algorithm preserves the positivity of $f^{(t)}$, {\it viz.} $f^{(t+1)}$ is guaranteed to be in $\mathbb{V}$ provided $f^{(t)} \in \mathbb{V}$ and $g \in \mathbb{V}$. Somewhat less trivial, it can also be shown that the algorithm preserves the mean values of the intermediate solutions, i.e. $\sum_{n,m} g_{n,m} = \sum_{n,m} f_{n,m}^{(t)}$ at each iteration $t$~\cite{Holmes92}. All these properties will play an important role in the proposed method as detailed below.

\section{MAP Formulation}
The approach proposed in this paper is based on the assumption that the original image $f$ in (\ref{model}) can be {\em sparsely represented} in the domain of a linear transformation. In particular, given an overcomplete and dense set $\{\phi_k\}_{k \in \mathcal{I}}$ of vectors in $\mathbb{R}^{N\times M}$, we define the synthesis operator ${\bf \Phi}$ to be given by
\begin{equation}\label{PHI}
{\bf \Phi}: \ell_2(\mathcal{I}) \rightarrow \mathbb{V}: c \mapsto {\bf \Phi}\{c\} = \sum_{k \in \mathcal{I}} c_k \phi_k,  
\end{equation} 
where $\mathcal{I}$ denotes the set of indices of the representation coefficients $c$, with $NM < \# \mathcal{I} < \infty$. Using this definition of $\bf \Phi$, one can rewrite the model equation (\ref{model}) as
\begin{equation}\label{newmodel}
g = \mathcal{P}\{ \Aa\{c\} \},
\end{equation}
with $\Aa$ being a composition of $\mathcal{H}$ and ${\bf \Phi}$, i.e. $\Aa := \Hh \circ {\bf \Phi}$, and $c \in \ell_2(\mathcal{I})$ being the representation coefficients of $f$ in the domain of $\bf \Phi$. In this new format, the original problem of reconstruction of $f$ is replaced by the problem of estimating its representation coefficients $c$.

Needless to say, due to the over-completeness of $\bf \Phi$, the definition of $c$ is not unique. This difficultly, however, can be easily overcome by requiring the representation coefficients $c$ to be as sparse as possible. A possible way to quantify the sparseness of $c$ is by measuring its $\ell_1$ norm $\|c\|_1$ \cite{Chen01}. It is worthwhile noting that this measure of sparseness has been successfully used in numerous fields of science, which include signal modeling \cite{Chen01}, compressed sensing \cite{Donoho06, Candes06}, independent component analysis and blind source separation \cite{Bofill01, Georgiev05}, inverse problems \cite{Daubechies03, Figueiredo03}, signal and image de-noising \cite{Donoho95, Aharon06}, morphological component analysis \cite{Starck04} together with its earlier version introduced in \cite{Michailovich02}.

Using the assumptions above, the problem of recovering the original image $f$ can be cast into the framework of MAP estimation, in which case the optimal $\hat{f}_{MAP}$ is found as a solution of the following maximization problem
\begin{equation}\label{MAP}
\hat{f}_{MAP} = \argmax_{c} \left\{ P(g \mid c) \, P(c) \right\},
\end{equation}
where $P(g \mid c)$ and $P(c)$ denote the likelihood function and the prior probability of $c$, respectively. 

Under the assumption of statistical independence, the likelihood $P(g \mid c)$ has the form of
\begin{equation}\label{like}
P(g \mid c) = \prod_{n=0}^{N-1} \prod_{m=0}^{M-1} \frac{ e^{- (\Aa\{c\})_{n,m}} (\Aa\{c\})_{n,m}^{g_{n,m}}}{g_{n,m} !}
\end{equation}
One the other hand, congruent to the assumption on minimality of $\|c\|_1$, the representation coefficients are assumed to be {\it i.i.d.} random variables that obey a zero-mean Laplacian distribution, {\it viz}
\begin{equation}\label{Laplace}
P(c) = \prod_{k \in \mathcal{I}} \frac{1}{2 \gamma} e^{-|c_k|/\gamma},
\end{equation}
where $\gamma>0$ is a scale parameter of the distribution.

Using the definitions (\ref{like}) and (\ref{Laplace}) and applying the standard log-transformation and the negation of sign to (\ref{MAP}), it is straightforward to show that the MAP estimation problem is equivalent to
\begin{equation}\label{costa}
\hat{f}_{MAP} = \argmin_c \{ E(c) \},
\end{equation}
\begin{equation}\label{costb}
E(c) = \langle {\bf 1}, \Aa\{c\} \rangle - \langle g, \log(\Aa\{c\})  \rangle + \lambda \|c\|_1 ,
\end{equation}
with $\lambda \equiv 1/\gamma$ being a regularization parameter.

Care should be exercised when specifying the domain of definition of $E(c)$ in (\ref{costb}). Indeed, the likelihood model of (\ref{like}) interprets the value $f_{n,m}$ as the mean of the corresponding random observation $g_{n,m}$. Since, in the case of a Poisson distribution, its first and second moments are equal, the values $f_{n,m}$ should be assumed to be non-negative (in accordance with our earlier assumption in (\ref{sspace})). Thus, the domain of the objective $E(c)$ is formally given by
\begin{equation}\label{dom}
{\rm \bf dom} \, E = \left\{ c \in \ell_2(\mathcal{I}) \mid ({\bf \Phi}\{c\})_{n,m} \geq 0 , \forall n,m \right\},
\end{equation} 
which is a convex set. Moreover, as long as $g_{n,m} > 0$ and the convolution operator $\Hh$ is non-degenerate (albeit, possibly, ill-conditioned), the objective functional $E(c)$ is guaranteed to be strictly convex. In this case, $E(c)$ admits a unique minimizer in ${\rm \bf dom} \, E$, which can be found by any optimization algorithm which is guaranteed to converge to a stationary point of $E(c)$ \cite{Boyd04}. Unfortunately, when either $g$ is not strictly positive or $\Hh$ possesses a non-trivial null-space, the convexity of $E(c)$ is not strict, and, as a result, the existence and uniqueness of its global minimizer cannot be a priori guaranteed. 

\section{Solution via Fixed-Point Iterations}
Numerous algorithms approaches can be used to minimize the cost functional in (\ref{costb}). In this paper, we propose a different method based on fixed-point iterations. To this end, we first notice that the first-order optimality condition for (\ref{costb}) has the following form
\begin{equation}\label{firstorder}
\nabla E(c = c^\ast) = \Aa^\ast\{{\bf 1}\} - \Aa^\ast \left\{ \frac{g}{\Aa\{c^\ast\}} \right\} + \lambda \, {\rm sign} (c^\ast) = 0,
\end{equation}
where $\Aa^\ast$ denotes the adjoint of $\Aa$ and $c^\ast$ is an extremum of $E(c)$. Note that, in (\ref{firstorder}), the non-differentiability of the absolute value at zero is resolved by letting $(|x|)' \big|_{x=0} = 0$. As will be shown below, the latter assumption is rather technical, as it has no impact on the proposed solution. 

Finally, by rearranging the terms in (\ref{firstorder}), one obtains
\begin{equation}
\Aa^\ast \left\{ \frac{g}{\Aa\{c^\ast\}} \right\} = \Aa^\ast\{{\bf 1}\}  + \lambda \, {\rm sign} (c^\ast),
\end{equation}
which, in turn, suggests the following fixed-point iteration algorithm
\begin{equation}\label{fixedpoint}
c^{(t+1)} = \Aa^\ast \left\{ \frac{g}{\Aa\{c^{(t)}\}} \right\} \frac{c^{(t)}}{\Aa^\ast\{{\bf 1}\}  + \lambda \, {\rm sign} (c^{(t)})}.
\end{equation}

The iterative procedure of (\ref{fixedpoint}) is multiplicative in nature -- the fact that has a number of important implications. First, the zero entries of $c$ are preserved by the iteration procedure, which means that if $c_i^{(t)} = 0$ for some $i \in \mathcal{I}$, then $c_i^{(t')} = 0$ for all $t' \ge t$. In this respect, letting $(|x|)' = {\rm sign}(x)$ seems to be a reasonable simplification, since the value of the gradient of $\|c\|_1$ at zero does not seem to have any influence on the result of the iterative procedure (\ref{fixedpoint}). Second, if the values of $c$ were allowed to be of an arbitrary sign, it would be generally impossible to guarantee a {\em monotone} convergence of $c^{(t)}$. To overcome this deficiency, in what follows the representation coefficients $c$ will be assumed to be non-negative. Moreover, if the representation vectors $\{\phi_i\}_{i \in \mathcal{I}}$ are constrained to be positive-valued as well, it is straightforward to verify that the iterations in (\ref{fixedpoint}) will preserve the positiveness of $c$. It is important to note that the above constraints on the sign of $c$ and $\{\phi_i\}_{i \in \mathcal{I}}$ should not be seen as a limitation, since there exists a body of works (see, for example,~\cite{Aharon05}), which describe the construction of positive-valued representation sets that allows the natural scenes to be represented in terms of both sparse {\em and} positive coefficients. 

Provided the elements of $\{\phi_i\}_{i \in \mathcal{I}}$ are positive values, the vector $v: = \Aa^\ast\{{\bf 1}\} = \left\{ \sum_{n,m} (\phi_i)_{n,m} \right\}$ will be positive valued as well, i.e. $v_i \ge 0, \forall i$. In this case, (\ref{fixedpoint}) can be rewritten in a more compact form as
\begin{equation}\label{SRL}
c^{(t+1)} = \Aa^\ast \left\{ \frac{g}{\Aa\{c^{(t)}\}} \right\} \frac{c^{(t)}}{v  + \lambda}.
\end{equation}
The iterations in (\ref{SRL}) can be initialized with a constant coefficient vector $c^{(0)}$ (e.g. $c_i^{(0)} = 1, \forall i$) and executed until the relative change $\| c^{(t+1)} - c^{(t)} \|_2 / \| c^{(t)} \|_2$ drops below a predefined threshold $0 < \epsilon \ll 1$. The above algorithm will be referred below to as the {\em sparse RL} method, or simply SRL. 

Comparing the recursions in (\ref{SRL}) and (\ref{RL}), one cannot avoid noting how similar the SRL and RL algorithms are. Indeed, replacing $c$ and $\Aa$ in (\ref{SRL}) by $f$ and $\Hh$, respectively, yields an update equation identical to that in (\ref{RL}) (up to the element-wise division by $v  + \lambda$ in (\ref{SRL})). Moreover, the assumption on the non-negativeness of $c$ and $\Aa$ made by SRL is parallel to the non-negativeness of $f$ and $\Hh$ exploited by the RL method. Further similarity between RL and SRL reveals itself in the objective functions of the two reconstruction methods. Specifically, the objective function in (\ref{costb}) can be rewritten as
\begin{align}\label{costbb}
E(c) &= \langle \Aa^{\ast}\{{\bf 1}\}, c \rangle - \langle g, \log(\Aa\{c\})  \rangle + \lambda \|c\|_1 = \nonumber \\
&= \langle v, c \rangle - \langle g, \log(\Aa\{c\})  \rangle + \lambda \|c\|_1 =  \|c\|_{w,1} - \langle g, \log(\Aa\{c\})  \rangle,
\end{align}
where $w = v + \lambda$ and $\|c\|_{w,1} = \sum_{i \in \mathcal{I}} w_i c_i$ is the $w$-weighted $\ell_1$-norm of the non-negative representation coefficients $c$. Thus, one can see that under the substitution of $c$ by $f$, $\Aa$ by $\Hh$ and $w=1$, the objective functionals (\ref{costbb}) and (\ref{CostL1}) are identical. Yet, while minimizing (\ref{CostL1}) forces the reconstruction of $f$ to have a sparse appearance, minimizing (\ref{costbb}) has the same effect on the representation coefficients. As opposed to the former, the latter case allows one to recover much broader classes of practical images, as it is demonstrated by the experimental results below. 

\section{Experimental Results}
\subsection{Reference methods and comparison measures}
Both one- and two-dimensional data sets have been used to test the performance of the proposed SRL algorithm. In the case of 1-D data, the SRL algorithm was compared against the RL method, while in the case of 2-D data, the comparison was made against the method described in \cite{Dey06}. Note that, just like SRL, the reference method of \cite{Dey06} belongs to the class of Bayesian estimators. Specifically, the method recovers the original image $f$ as a maximizer of its posterior probability, computed under the model of (\ref{model}) and the {\em total-variation} (TV) prior. The maximization can be performed iteratively according to the following update equation
\begin{equation}\label{RLTV}
f_{t+1} =   \frac{f_t}{1 - \gamma \, {\rm div} \left( \nabla f_t \big/ \|\nabla f_t\| \right) } \ {\bf H}^\ast \left\{ \frac{g}{ {\bf H}\{f_t\} }  \right\},
\end{equation}
where $\gamma > 0$ is a regularization parameter, which has been set to be equal to 0.002 as prescribed in \cite{Dey06}. For the convenience of referencing, the above algorithm will be referred below to as the RLTV method.

To quantitatively compare between the performance of various reconstruction algorithms, two comparison measures have been used. The first, {\em normalized mean-squared error} (NMSE), is defined as follows. Let $f$ be an original image and $\hat{f}$ be an estimate of $f$. Then, the NMSE can be defined as
\begin{equation}\label{NMSE}
{\rm NMSE} = \mathcal{E} \left\{ \frac{\| f - \hat{f} \|_F^2}{\| f \|_F^2} \right\},
\end{equation}
with $\| \cdot \|_F$ being the Frobenius matrix norm, and $\mathcal{E}$ being the operator of expectation. In the current study, the expectation was approximated by its corresponding sample mean based on the results of 200 independent trials.

It has been recently argued that the NMSE may not be an optimal comparison measure as long as human visual perception is concerned. For this reason, the structural similarity (SSIM) index proposed by \cite{Wang04} was employed as well as an alternative performance metric.

\subsection{One-dimensional Reconstruction}
The first part of our experimental study is concerned with the recovery of piece-wise constant signals of length $N$. For this case, a natural choice would be to define the basis functions $\{\phi_i\}_{i \in \mathcal{I}}$ to be scaled and spatially shifted versions of a rectangular (Haar) box. Specifically, let $j = 0, 1, \ldots, \lfloor \log_2(N) \rfloor - 1$ be a (dyadic) resolution index, and $\phi_j \in \mathbb{R}^N$ be defined as
\begin{equation}\label{box}
\phi_j[n] = 
\begin{cases}
2^{-j/2}, &\mbox{ for } 0 \le n < 2^j - 1 \\
0,          &\mbox{ for } 2^j \le n < N,
\end{cases}
\end{equation}
where the normalization factor $2^{-j/2}$ is used to guarantee $\|\phi_j\|_2 = 1$ for all $j$. Also, let 
\begin{equation}\label{shift}
\mathcal{Z}^k: \mathbb{R}^N \rightarrow \mathbb{R}^N: x[n] \mapsto x[(n-k) \, {\rm mod} \, N]
\end{equation}
be the operator of (causal) circular shift by $k$ points. Obviously, for each $\phi_j$ there are a total of $N$ non-repetitive shifts. Consistent with the notations standard in wavelet theory, let $\phi_{k,j} : = \mathcal{Z}^k\{\phi_j\}$ denote the function $\phi_j$ (circularly) shifted by $k$ points to the right (in which case $\phi_{k,j}$ can be viewed as a rectangular box function supported on $[(k, \, k+2^j-1) \, {\rm mod} \, N]$). Moreover, let ${\bf \Phi} := \{\phi_{k,j}\}$ be a matrix of size $N \times N \lfloor \log_2(N) \rfloor$, whose columns are formed by functions $\phi_{k,j}$ with $k=0,1,\ldots, N-1$ and $j=0,1,\ldots, J-1$. In this case, given an arbitrary vector of representation coefficients $c$ of size $N \lfloor \log_2(N) \rfloor \times 1$, its corresponding signal $f$ can be synthesized as $f = {\bf \Phi} \, c$.  

\begin{figure}[htp]
\centering
\includegraphics[width=4.5in]{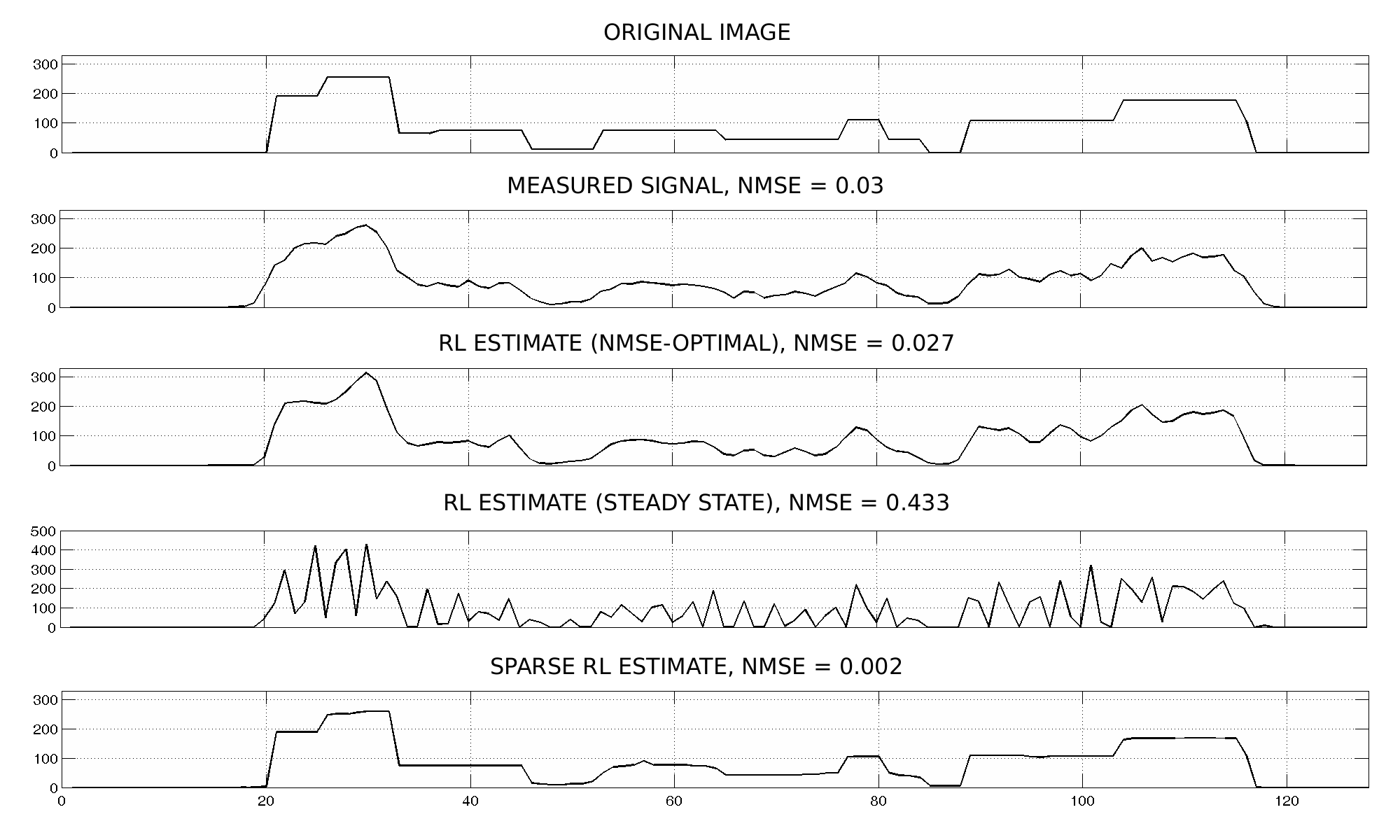} 
\caption{High-count scenario (from up to bottom): The original signal, its blurred version, the blurred signal contaminated by Poisson noise, the MSE optimal RL estimate, the steady-state RL estimate, and the SRL estimate.}
\label{Fig1}
\end{figure}

\begin{figure}[htp]
\centering
\includegraphics[width=4.5in]{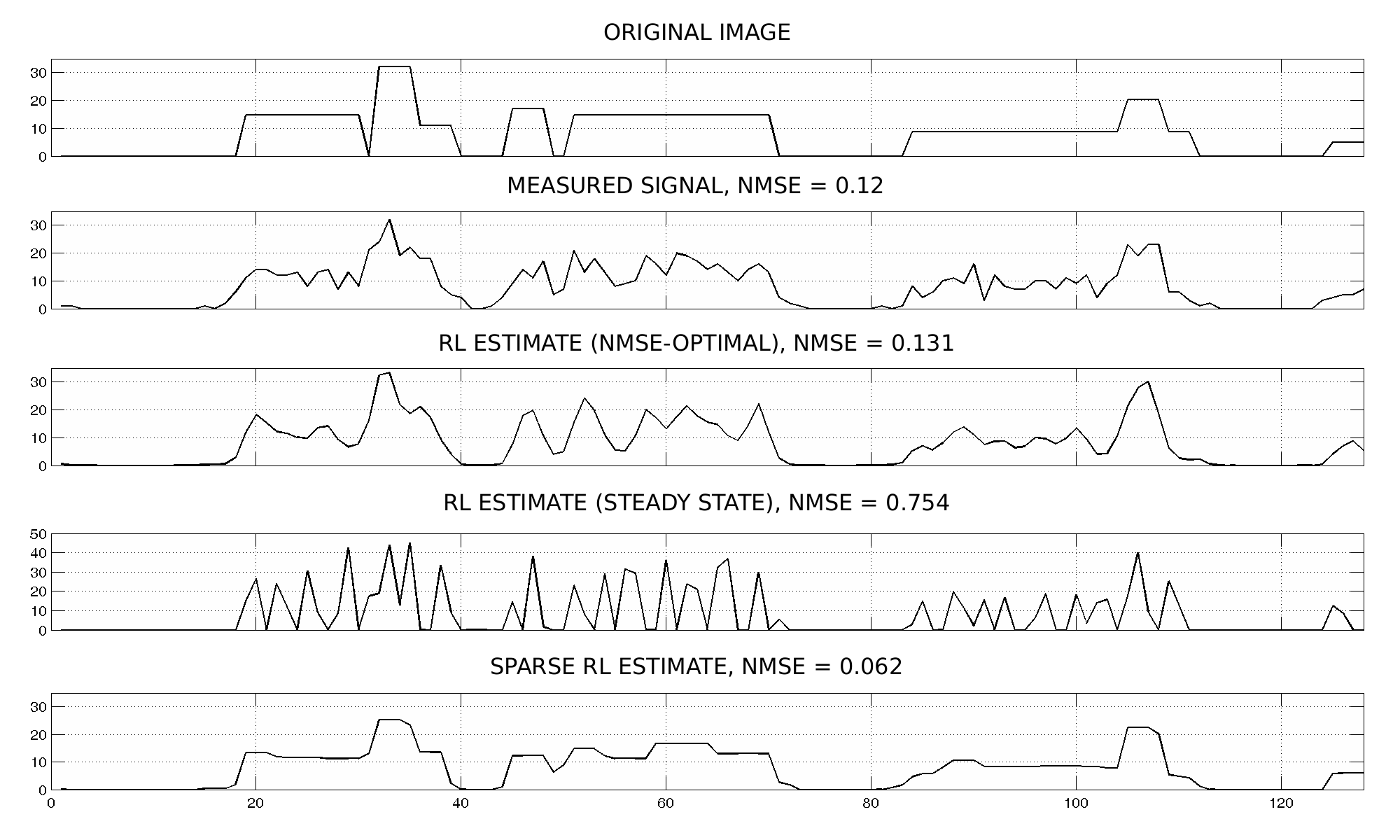}
\caption{Low-count scenario (from up to bottom): The original signal, its blurred version, the blurred signal contaminated by Poisson noise, the MSE optimal RL estimate, the steady-state RL estimate, and the SRL estimate.}
\label{Fig2}
\end{figure}

The {\em dictionary} $\bf{\Phi}$ constructed above is severely overcomplete, which has the disadvantage of high computational complexity associated with the computations of ${\bf \Phi} \, c$ and ${\bf \Phi}^T f$. To improve the computational efficiency, the index $j$ was restricted to four resolution levels, viz. $j=2,3,4$, and $5$. As a result, the size of $\bf \Phi$ was equal to $128 \times 512$ for $N=128$. The coefficients $c$ used for the synthesis of test signals had about $1.5 \div 3$ \% of non-zero entries, drawn from a uniform distribution. The blurring artifact was simulated by convolving the test images with a Gaussian kernel $\Hh$ whose -3 dB cut-off frequency was set to be equal to $0.2 \pi$. The maxima of the resulting signals were set to two different values, namely 256 and 32, in order to simulate high- and low-count detection scenarios, respectively. As a final step, the blurred signals were contaminated by Poisson noise (see the second subplots in Fig.~\ref{Fig1} and Fig.~\ref{Fig2} for typical examples of data signals). 

Next, the RL and SRL algorithms were applied to the synthesized data signals with $\lambda = 0.2$. Unfortunately, {\em no} monotonous convergence in NMSE could be achieved in the case of the RL algorithm. The steady-sate estimates obtained using this classical method were unacceptably noisy (see the fourth subplots of Fig.~\ref{Fig1} and Fig.~\ref{Fig2} for typical examples of the RL estimates).  For this reason, it was decided to terminate the RL estimations before the steady-state was reached, at the point when the NMSE value was minimized. (Note that such NMSE-optimal RL estimates are impossible to compute in real-life scenarios, as their computation requires knowing the original images.) The SRL algorithm, on the other hand, produced monotonous convergence in NMSE, producing substantially smaller values of NMSE as compared to the case of RL estimation. For the case of high- and low-count scenarios, some typical reconstruction results are depicted in Fig.~\ref{Fig1} and Fig.~\ref{Fig2}, respectively, from which the superiority of the proposed method can be easily appreciated. It is also interesting to note that, as predicted by the theory, the RL algorithm attempts to arrive at a sparse estimate of the signal of interest, which contradicts the piece-wise smooth nature of our test signals. The SRL method, on the other hand, require the sparsity of the representation coefficients, which appears to be a much more realistic assumption for the case at hand.  

\begin{figure}[htp]
\centering
\includegraphics[width=5in]{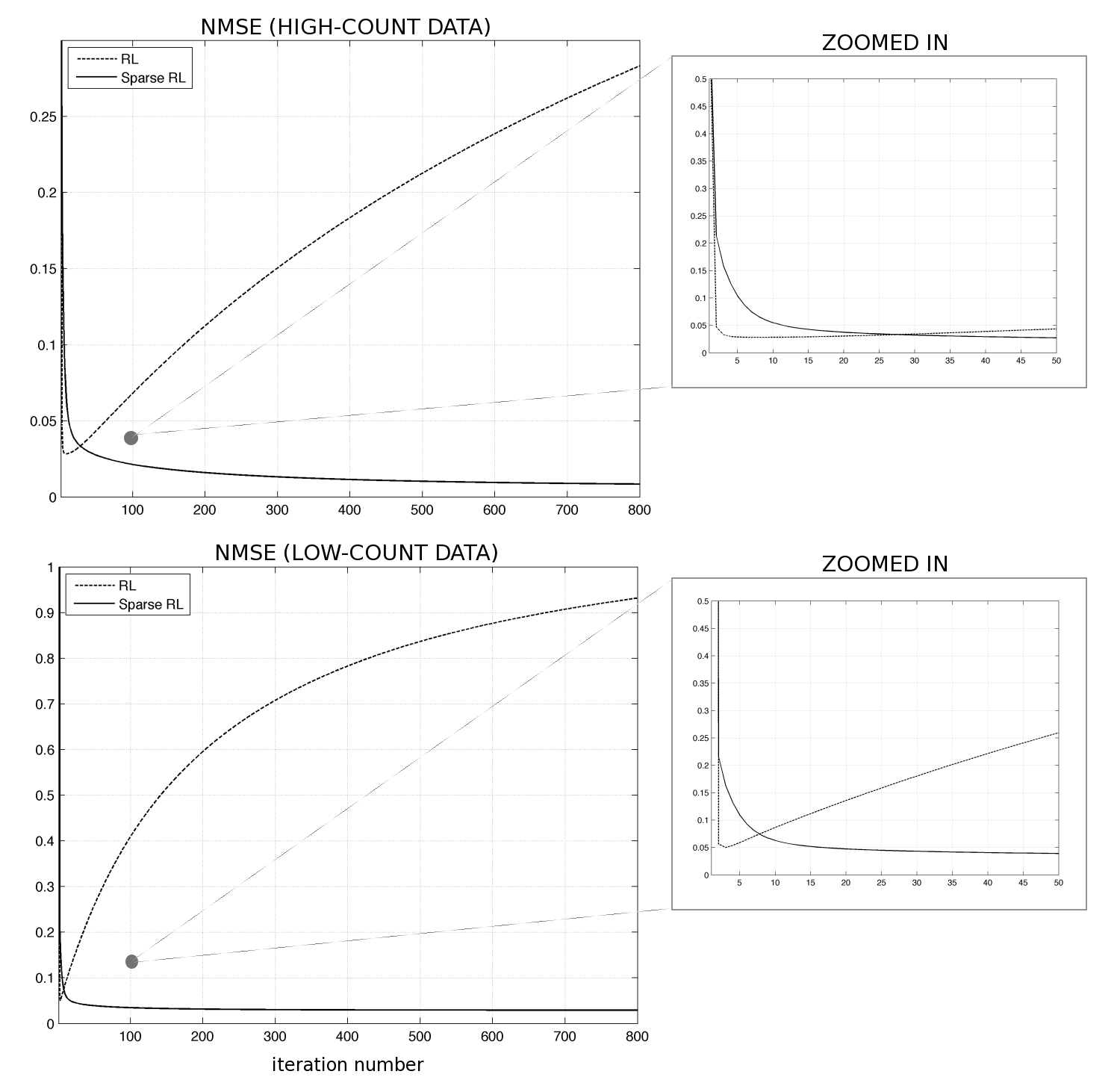} 
\caption{The convergence of NMSE values produced by the RL and SRL algorithms in the case of high (upper subplot) and low (lower subplot) count scenarios.}
\label{Fig3}
\end{figure}

A quantitative comparison of the two reconstruction algorithms is presented in Fig.~\ref{Fig3}, which depicts the values of NMSE produced by the RL and SRL methods as a function of the number of iterations. It is important to emphasize that each value of the NMSE in Fig.~\ref{Fig3} is a result of averaging the errors obtained in a series of independent trials, where both the true signals and noises were drawn randomly. Observing Fig.~\ref{Fig3}, one can see that SRL results in considerably lower values of the NMSE as compared to the case of RL. As well, it converges {\em monotonously} to a steady-state solution after a relatively small number of iterations. On the other hand, the convergence of RL is not monotone -- the fact that necessitates termination of the algorithm after a predefined number of iterations. Needless to say, determining a proper number of iterations is a data-dependent and somewhat esoteric task in practice.

\subsection{Two-dimensional Reconstruction}
In the second part of the experimental study, the SRL algorithm is examined using imaging data. As reference methods, the standard RL method as well as the RLTV algorithm (as specified by (\ref{RLTV})) are employed to assess the performance of SRL in a comparative way. Note that as neither RL nor RLTV could provide acceptable steady-state results, their iterations needed to be terminated at an earlier point. In particular, the iterations were terminated at the point when the NMSE produced by the algorithms reached a minimum value. It is important to reiterate that such ``MSE-optimal" termination rule is not realizable in real-life scenarios where the original images are unknown. Hence, the RL and RLTV estimates demonstrated below represent the best possible result which could be achieved using these reconstruction methods. 

Implementation of the SRL algorithm relies on the availability of a dictionary of positive-valued atoms $\{\phi_i\}_{i \in \mathcal{I}}$. In the 2-D experiments, we explore two possible approaches to the definition of such a dictionary, viz. unsupervised and supervised. In the first case, the dictionary is composed of shift-invariant subsets generated by cubic B-splines \cite{Unser99}, and therefore it does not depend on the nature of imagery data at hand. In the second case, the dictionary is learned from a training set that is supposed to represent the family of objects which the measured image is likely to belong to.

\subsubsection{Unsupervised dictionary} For the sake of reproducibility of the results of this paper,  we provide some necessary details on the construction of the cubic spline dictionary. To this end, let ${\bf 1}_{2^j\times1}$ denote a vector of length $2^j$, whose elements are all equal to 1. Also, let $b_3 = [1, \, 4, \, 1]/6$ be the vector of integral values of the cubic B-spline and $b_j$ be the vector of integral values of the cubic B-spline scaled by a dyadic factor of $2^j$, with $j=0,1,\ldots, J-1$. The vectors $b_j$ have the length of $M_j = 2^{j+2} - 1$ points and they can be computed recursively according to \cite{Unser99}
\begin{equation}
b_j = 2^{-3 j} \big( \underbrace{{\bf 1}_{2^j\times1} \ast \ldots \ast {\bf 1}_{2^j\times1}}_{\mbox{\scriptsize 4 times}} \big) \ast b_3,
\end{equation}
where $\ast$ denotes the operation of convolution. As prescribed by common practice, the vectors  $b_j$ can be subsequently normalized to obey $\|b_j\|_2 = 1$.

The 1-D kernels $b_j$ can be extended into isotropic and separable 2-D (discrete) splines $B_{j}$ by means of the tensor product, i.e. $B_{j}[n,m] = b_{j}[n] \,b_{j}[m]$, $0 < j \le J-1$. Consequently, the true image $f$ is assumed to belong to the space of all integer translations of the discrete kernels $B_j$. Under periodic boundary conditions, there are a total of $N\times M$ representation coefficients  $c_j \in \mathbb{R}^{N\times M}_+$ for each $j=0,1,\ldots, J-1$. Let the array of all these coefficients be denoted by $c = \{c_j\}_{j=0}^{J-1}$. Then, the operator $\Aa$ can be shown to be given by
\begin{equation}\label{Adir}
\Aa: \mathbb{R}^{N\times M\times J}_+ \rightarrow \mathbb{R}^{N\times M}_+: c \mapsto f= \Hh\Big\{  \sum_{j=0}^{J-1} c_j \circledast B_j  \Big\},
\end{equation}
while its adjoint $\Aa^\ast$ is given by
\begin{equation}\label{Aadj}
\Aa^\ast: \mathbb{R}^{N\times M}_+ \rightarrow \mathbb{R}^{N\times M\times J}_+: f \mapsto c = \Big\{  \Hh\{f\} \circledast B_j  \Big\}_{j=0}^{J-1},
\end{equation}
with $\circledast$ standing for circular convolution\footnote{For some examples of related MATLAB$\textregistered$ codes and their use visit {\tt www.ece.uwaterloo.ca/}\~{\tt olegm/research.html}.}. According to (\ref{Aadj}), an $N\times M$ image produces a total of $J\,N\,M$ spline coefficients, and hence the overall complexity of applying $\Aa$ and $\Aa^\ast$ is comparable to that of a stationary wavelet transform.

\subsubsection{Supervised dictionary} In the supervised case, the dictionary $\{\phi_i\}_{i \in \mathcal{I}}$ is designed adaptively based on a set of training examples that represent the family of images, to which $f$ is expected to belong. Consequently, the training procedure aims to find a set of positive-valued representation functions, using which one can represent the training images with a minimum possible number of representation coefficients. In this work, the training was performed by means of the the non-negative kSVD (NN-kSVD) algorithm\footnote{A software for the NN-kSVD algorithm can be obtained from {\tt http://www.cs.technion.ac.il/}\~{\tt elad/software/}.} proposed in \cite{Aharon05}. In compliance with the Sparseland model of \cite{Aharon06}, the training was applied to $16\times 16$ overlapping segments of a total of 11 MRI scans of the brain, which did not include the scans used in reconstruction. Each of these segments was represented by a linear combination of a total of 512 atoms, thereby resulting in the over-completeness factor of 2 per segment. For the formal definition of operators $\Aa$ and $\Aa^\ast$ the reader is kindly referreded to \cite{Aharon06}.    

\subsubsection{Reconstruction of MRI scans} 
\begin{figure}[htp]
\centering
\includegraphics[width=5in]{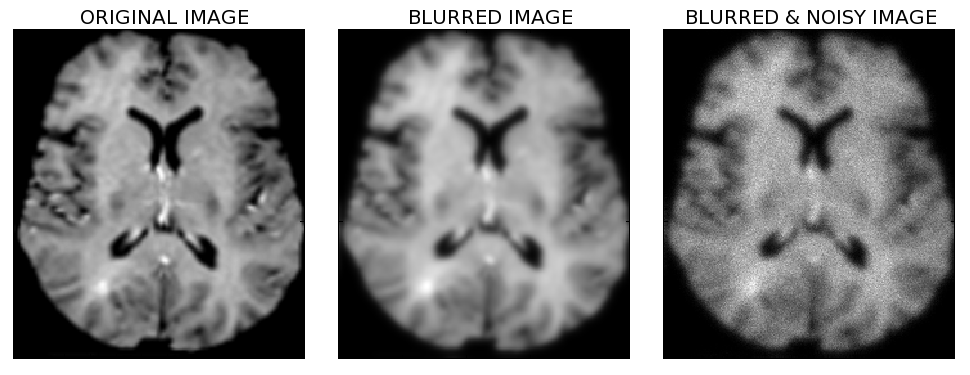}
\caption{(Left subplot) Original image of a cross section of the brain; (Middle subplot) Blurred image; (Right subplot) Blurred and noisy image.}
\label{Fig4}
\end{figure}

The blurring kernel used in our next experiment was defined to be $h[i,j] = (i^2+j^2+1)^{-1}$, with $i,j = -7,\ldots,7$ \cite{Elad07}, while the regularization parameter $\lambda$ was set to be equal to 0.1 in both unsupervised and supervised cases. A typical example of the original image used in this study along with its blurred and noise-contaminated versions are shown in the leftmost, middle, and the rightmost subplots of Fig.~\ref{Fig4}, respectively. The ``measured" MRI scans have been obtained by contaminating the blurred image with Poisson noise giving rise to an SNR of approximately 15 dB. 

\begin{figure}[htp]
\centering
\includegraphics[width=4.5in]{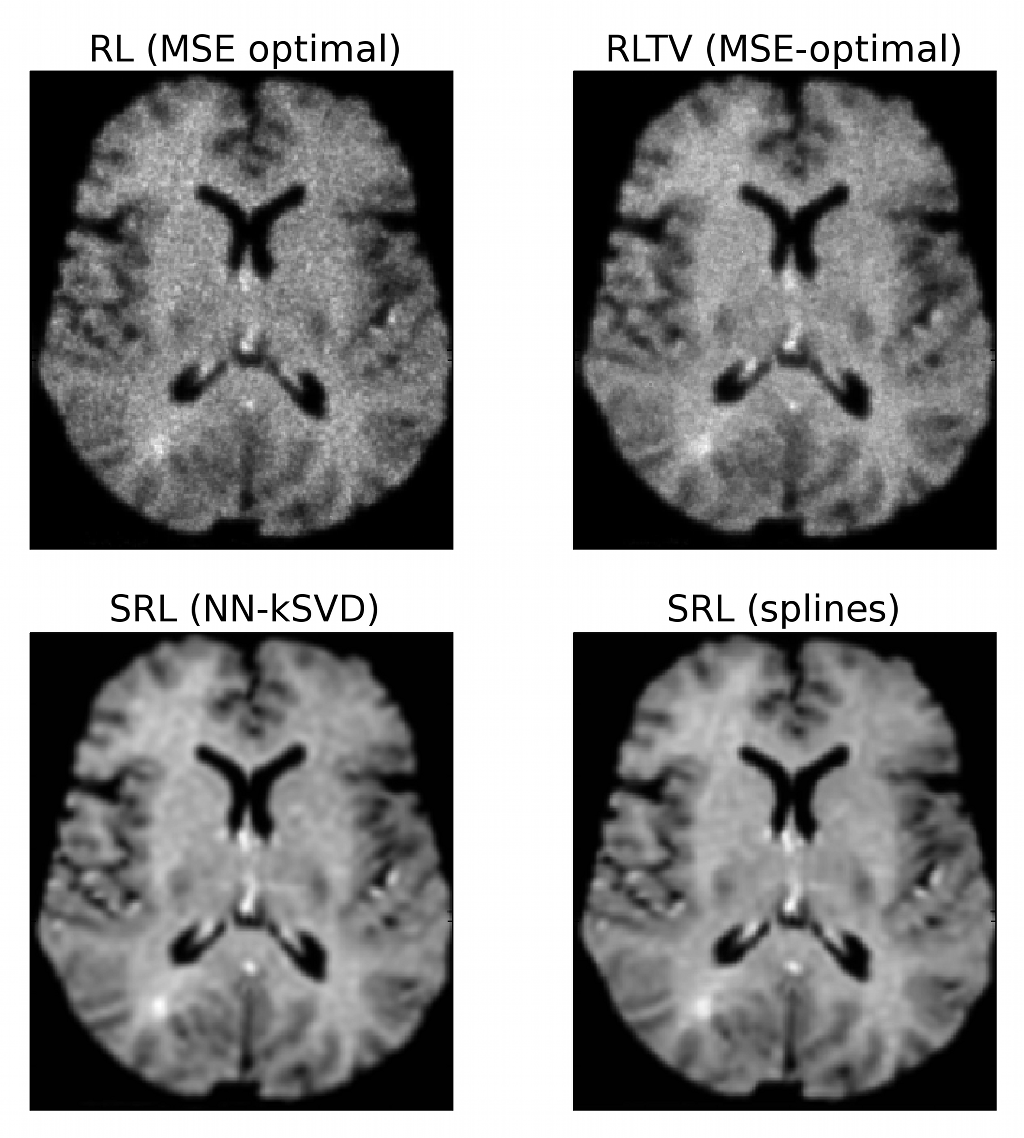}
\caption{Image reconstruction results corresponding to Fig.\ref{Fig4}: (Upper row of subplots) The MSE-optimal RL and RLTV estimates; (Lower row of subplots) supervised (kSVD) and unsupervised (splines) SRL recovery.}
\label{Fig5}
\end{figure}

Some typical reconstruction results produced by the proposed and reference methods are shown in Fig.~\ref{Fig5}. One can see that the SRL method is capable of better recovering the details of the true image as compared to the alternative approaches. It can also be seen that the SRL estimates possess higher resolution and contrast as compared to the alternative approaches. These conclusions are further supported by the quantitative measures of Table~\ref{T1}, which compares the NMSE and SSIM indices of different estimates. As shown by the table, the supervised SRL method produces the lowest NMSE and the largest SSIM index among all the methods under comparison.  

\begin{table}[here]
\caption{NMSE and SSIM values obtained using the reconstruction methods under comparison.}
\centering
\begin{tabular}{l*{4}{c}r}
\toprule[1.5pt]
& RL (MSE-optimal) & RLTV (MSE-optimal) & SRL (Splines) &  SRL (NN-kSVD) \\
\midrule
NMSE	&	0.022	&	0.011	&	0.007	& {\bf 0.003} \\
SIM  	& 	0.71		& 	0.77		&	0.89		& {\bf 0.91} \\
\bottomrule
\end{tabular}
\label{T1}
\end{table}

\section{Discussion and Conclusions}
In the current paper, a new regularized version of the RL algorithm has been presented, which is advantageous in a number of ways. First, the proposed method is general in its formulation. The latter allows applying the same reconstruction procedure to a number of different settings, such as image de-noising or image enhancement through deconvolution. Attuned to deal with Poissonian noises, the method can therefore be applied for the reconstruction of imagery data produced by many important image modalities, including optical, microscopic, turbulent, and nuclear imaging, just to name a few. Moreover, whilst many alternative reconstruction methods take advantage of certain simplifying assumptions about the noise nature, the proposed technique is optimized to deal with the realistic noise model at hand.

Another important advantage of the proposed method consists in the generality of the prior model used to describe the nature of true images. Specifically, the images have been assumed to admit a sparse representation in the domain of a properly chosen linear transform. Note that the above {\it a priori} model is nowadays considered to be a standard model in numerous applications of the theory of sparse representations. This is what allows the SRL algorithm to recover piecewise smooth images -- the task impossible to accomplish by means of the standard RL procedure for its property to favour {\em spatially} sparse reconstructions. 

Yet another critical advantage of the proposed SRL algorithm is in its algorithmic structure, which allows solving non-smooth optimization problems at the computational cost of a steepest descent procedure. Consequently, the computational load required by the proposed method is relatively small, which allows the method to be applied for the solution of large-scale problems and/or for processing of large data sets.

\end{document}